\pdfoutput=1
\documentclass[10pt,twocolumn,letterpaper]{article}
\usepackage[T1]{fontenc}
\usepackage[pagenumbers]{cvpr}
\usepackage{graphicx}
\usepackage{amsmath,amssymb,amsfonts}
\usepackage{booktabs}
\usepackage{multirow}
\usepackage{xcolor}
\definecolor{cvprblue}{rgb}{0.21,0.49,0.74}
\usepackage{subcaption}
\usepackage[pagebackref,breaklinks,colorlinks,allcolors=cvprblue]{hyperref}
\usepackage{placeins}
\usepackage[capitalize]{cleveref}
\newcommand{\rose}{RoSeDiT}
\title{Role-Separated Diffusion Transformers:
Global Rule Isolation and Deterministic Sampling Manifold Compression}
\author{Zhiguang Liu \qquad Yi Shang\\
Department of Electrical Engineering and Computer Science, University of Missouri\\
{\tt\small lz7fd@missouri.edu \quad shangy@missouri.edu}
}
\makeatletter
\g@addto@macro\@maketitle{%
  \vspace{-10mm}
  \begin{center}
  \includegraphics[width=0.9\textwidth]{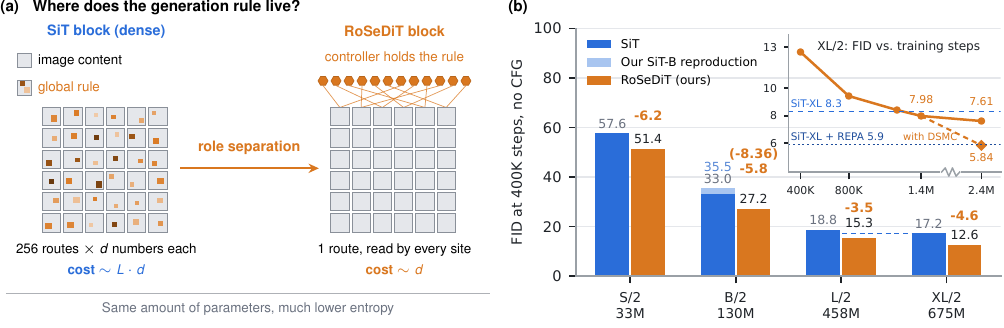}
  \end{center}
  \vspace{-5mm}
  \captionof{figure}{\textbf{Global rule isolation and generation performance.} (a) Illustration of role separation. SiT applies global conditioning through shared modulation across image tokens, whereas \rose{} maintains dedicated controller representations that dynamically inform image-token modulation, motivated by reducing the representational complexity of global transformation rules. (b) FID comparison at 400K steps at the same level of parameter count: \rose{} consistently outperforms SiT across all four model scales, with \rose{}-L/2 surpassing SiT-XL/2 by 1.9 FID points. \rose{}-XL/2 achieves an FID of 7.98 at 1.4M steps, outperforming SiT-XL/2's 8.3 at 7M steps. With DSMC, \rose{}-XL/2 reaches 5.84 at 2.4M steps, compared with 7.61 without DSMC, and surpasses SiT + REPA's 5.9 at 4M steps. }
  \label{fig:teaser}
  \vspace{1.5mm}
}
\makeatother
\begin{document}
\maketitle
\begin{abstract}
We present Role-Separated Diffusion Transformers (RoSeDiT), a new diffusion transformer architecture that isolates the global rule from the image tokens. RoSeViT showed in visual reasoning that a vanilla ViT spreads a shared rule over all $L$ tokens at an entropy of order $Ld$, whereas an enforced controller route reduces it to order $d$. In class-conditional DiT and SiT, the class and the noise level are explicitly conditioned by adaLN, so a RoSeViT-style controller channel on the ViT side is insufficient to establish the full role separation. By introducing the adaLN listening mechanism, which makes adaLN compute under the controller's command, RoSeDiT brings the entropy theorem and role separation into the latent diffusion domain. Under the same SiT training and evaluation protocol at the same level of parameter count, RoSeDiT surpasses SiT at all four model scales at 400K steps, and RoSeDiT-L/2 even surpasses SiT-XL/2's 17.2 FID with 15.34 and 32\% fewer parameters. RoSeDiT-XL/2 surpasses SiT-XL/2's 7M-step FID of 8.3 at 1.4M steps with 7.98, using one-fifth as many training steps. We further observe that models with the same denoising loss can have a large FID gap, which indicates a misalignment between the denoising objective and the image manifold. We formalize this phenomenon into Deterministic Sampling Manifold Compression (DSMC), a two-stage training schedule with which RoSeDiT-XL/2 surpasses SiT + REPA's 4M-step FID of 5.9 at 2.4M steps with 5.84.
\end{abstract}

\section{Introduction}
\label{sec:intro}

Diffusion transformers have emerged as a leading architecture for image generation. DiT~\cite{peebles2023dit} replaced U-Net-based denoisers with transformers operating on latent image patches, while SiT~\cite{ma2024sit} improved generation quality through the stochastic interpolant framework. Subsequent advances have achieved substantial gains through representation alignment (REPA)~\cite{yu2025repa}, improved latent representations and autoencoders~\cite{yao2025vavae,zheng2025rae}, and advances in training objectives and sampling. However, DiT and SiT rely on adaptive layer normalization (adaLN) to modulate image tokens using fixed class and timestep embeddings. Although effective, this mechanism does not explicitly maintain an evolving representation of global conditioning information that adapts to intermediate image features. This motivates reconsidering how global conditioning is represented and applied within diffusion transformer blocks.

A related challenge arises in visual reasoning, where image transformations are governed by global rules shared across spatial locations. RoSeViT~\cite{liu2026rosevit} investigated the representational complexity of such transformations, showing that site-wise rule representations can incur entropy scaling of $\Theta(Ld)$, where $L$ denotes the number of image tokens and $d$ the rule dimension. By separating global rules from image content through dedicated controller tokens and a structured information pathway, RoSeViT established a framework for reducing complexity to $\Theta(d)$ when the controller route is enforced. However, its structural enforcement mechanism introduces additional parameters, complicating parameter-matched comparisons. This motivates exploring whether role separation can improve diffusion transformer performance at the same level of parameter count.

In class-conditional DiT and SiT, the class and the noise level are explicitly conditioned by adaLN. Thus, adding an additional RoSeViT-style controller channel on the ViT side is insufficient to establish the full role separation. We present \rose{}, the Role-Separated Diffusion Transformer, which realizes global rule isolation through three components (\cref{fig:block}). Twelve controller tokens maintain an evolving representation of the global condition, a listening mechanism lets adaLN compute its modulation from their states, and a structural pass restricts each image token to its $3{\times}3$ neighborhood and the controller tokens. \rose{} maintains the same level of parameter count by reusing the dense attention weights.

We evaluate \rose{} on class-conditional ImageNet $256{\times}256$ generation under the SiT training and evaluation protocol. At 400K steps, \rose{} consistently outperforms SiT across four model scales, with larger relative improvements at greater scales. Notably, \rose{}-L/2 achieves an FID of 15.34 with 458M parameters, outperforming SiT-XL/2's 17.2 with 675M parameters. With continued training, \rose{}-XL/2 reaches an FID of 7.98 in 1.4M steps, surpassing SiT-XL/2's 8.3 after 7M steps without classifier-free guidance. We further introduce Deterministic Sampling Manifold Compression (DSMC), a two-stage schedule that selectively reuses fixed (image, noise level, noise) triples during training. With DSMC, \rose{}-XL/2 achieves an FID of 5.84 at 2.4M steps, compared with 7.61 without DSMC, and surpasses SiT + REPA's 5.9 after 4M steps. \rose{} is able to combine with classic SiT methods such as REPA. At the B scale, \rose{} + REPA reaches 21.0, a 3.4 FID lead over SiT + REPA's 24.4.

Our main contributions are summarized as follows:
\begin{itemize}
\setlength\itemsep{0pt}

\item \textbf{\rose{}: Role-Separated Diffusion Transformers.} We extend global rule isolation from visual reasoning to latent diffusion through dedicated controller tokens and a parameter-shared structural pass, establishing a distinct pathway for global conditioning at the same level of parameter count.

\item \textbf{The adaLN listening mechanism.} We introduce a mechanism that dynamically incorporates evolving controller states into image-token modulation, making role separation effective in diffusion transformers. Ablations identify listening as a major contributor to the performance improvements.

\item \textbf{Deterministic Sampling Manifold Compression (DSMC).} We propose a two-stage training schedule that combines deterministic sample reuse with fresh stochastic samples, substantially improving generation quality without modifying the model architecture.
\end{itemize}

\section{Related Work}
\label{sec:related}

\paragraph{Diffusion transformer architectures and conditioning.}
DiT~\cite{peebles2023dit} introduced transformer-based denoising over latent image patches and demonstrated the effectiveness of adaptive layer normalization with zero-initialized modulation (adaLN-Zero) for class-conditional generation. SiT~\cite{ma2024sit} retained the DiT architecture while adopting the stochastic interpolant framework~\cite{albergo2023interpolants,lipman2023flow,liu2023rectified}, improving generation quality across model scales. Subsequent designs include long skip connections in U-ViT~\cite{bao2023uvit}, joint text-image attention in MM-DiT~\cite{esser2024sd3}, and decoupled conditioning and prediction in DDT~\cite{wang2025ddt}, while Registers~\cite{darcet2024registers}, Perceiver~\cite{jaegle2021perceiver}, Swin Transformer~\cite{liu2021swin} and neighborhood attention~\cite{hassani2023nat} add auxiliary tokens or restricted attention to transformers. As an alternative to fixed conditioning vectors, Improved Mean Flows (iMF)~\cite{geng2025imf,geng2025meanflow} represents discrete and continuous conditions using in-context tokens. RoSeDiT instead couples evolving controller tokens to adaLN, so that conditioning adapts to intermediate image representations, and its structural pass runs alongside dense attention while reusing attention parameters.

\paragraph{Role separation in visual reasoning.}
RoSeViT~\cite{liu2026rosevit} introduced role separation for visual reasoning tasks involving global transformation rules shared across spatial locations. Building on the ARC benchmark~\cite{chollet2019measure} and the VARC framework~\cite{hu2026varc}, RoSeViT uses dedicated controller tokens to represent global rules while image tokens maintain spatial content. A masked structured pass promotes controller-mediated information exchange alongside standard dense attention. RoSeViT also developed a theoretical framework relating the representational complexity of site-wise transformation rules to the separation of global and local information, supported by empirical improvements over conventional ViTs. Unlike visual reasoning, where transformation rules must be inferred from examples, class-conditional diffusion provides global conditions explicitly. Thus the direct adoption of the controller route from visual reasoning didn't automatically bring the entropy advantage; the global conditioning needs to be isolated by a new adaLN listening mechanism.

\paragraph{Complementary advances in diffusion generation.}
Beyond backbone architecture, diffusion generation has benefited from improved training objectives, latent representations, and inference strategies. REPA~\cite{yu2025repa} accelerates diffusion transformer training by aligning intermediate features with representations from a pretrained visual encoder such as DINOv2~\cite{oquab2024dinov2}. REPA-E~\cite{leng2025repae}, VA-VAE~\cite{yao2025vavae}, and representation autoencoders~\cite{zheng2025rae} investigate improved representations and their integration with generative modeling. MeanFlow~\cite{geng2025meanflow} develops an objective for one-step generation, while classifier-free guidance~\cite{ho2022cfg} and EDM~\cite{karras2022edm} improve sampling strategies. These approaches address complementary aspects of generative modeling relative to RoSeDiT's controller-based conditioning architecture. In particular, we demonstrate that RoSeDiT can be combined with REPA to further improve generation quality (\cref{sec:repa}).

\paragraph{Deterministic Sampling Manifold Compression (DSMC).}
Prior work has explored diffusion training schedules, loss weighting, and sample assignment strategies~\cite{bengio2009curriculum,hang2023minsnr,tong2024otcfm,li2024immiscible}. DSMC takes a different approach by selectively reusing identical (image, noise level, noise) triples through a two-stage schedule that combines fixed samples with rotating pools of samples while retaining stochastic sampling. We initially observed this phenomenon when random seeds were inadvertently reused during resumed training, and subsequently investigated and formalized it as DSMC. To the best of our knowledge, this specific sample-reuse strategy has not been explored in prior diffusion training methods. We present the schedule in \cref{sec:dsmc} and evaluate it in \cref{sec:longer}.

\section{Theory of Role Separation}
\label{sec:background}

RoSeViT~\cite{liu2026rosevit} established a theoretical framework for separating global transformation rules from spatial representations. Under bi-Lipschitz assumptions, the metric entropy of independently represented site-wise rules scales as $\Theta(Ld\log(1/\varepsilon))$, where $L$ is the number of image tokens, $d$ is the rule dimension and $\varepsilon$ is the precision. Representing a shared rule through a $d$-dimensional controller instead reduces the entropy to $\Theta(d\log(1/\varepsilon))$ for the corresponding restricted function class. This motivates global rule isolation as a means of reducing representational complexity.

In class-conditional diffusion, the class label $y$ and noise level $t$ specify a global condition $r=(y,t)$ shared across image tokens. DiT and SiT apply this condition through adaLN modulation derived from fixed embeddings, whereas RoSeDiT maintains evolving controller states and enables modulation to read them dynamically. Together with the structural pass, this listening mechanism establishes a controller-mediated pathway for global conditioning while retaining dense attention. We extend RoSeViT's entropy perspective to this setting through the following theorem.

\textbf{Theorem 1 (Conditioned diffusion entropy).} Under RoSeViT's assumptions, let $r$ reach the $L$ image tokens through the modulation of every block, computed from one state $u$ shared by all image tokens.
Let $\mathcal{F}_{\mathrm{site}}$ be the site-wise family, in which each token may independently represent a $d$-dimensional rule, and $\mathcal{F}_{\mathrm{ctrl}}=\{F_u : u\in[0,1]^d\}$ the controller-based family, in which all tokens access the rule through $u$.
For a set $\mathcal{R}$ of such families available to an architecture, let $H_\varepsilon(\mathcal{R})=\min_{\mathcal{F}\in\mathcal{R}}\log N(\varepsilon,\mathcal{F},d_\infty)$ be its metric entropy, where $N$ is the covering number at precision $\varepsilon$ in the sup metric $d_\infty$.
(i) If $u=c$ reads only the condition, with $c=e_t(t)+e_y(y)$ the timestep and class embeddings, then for any controller tokens and mask
\[
\mathcal{R}_{\mathrm{dense}}=\{\mathcal{F}_{\mathrm{site}}\}, \qquad H_\varepsilon(\mathcal{R}_{\mathrm{dense}})=\Theta\big(Ld\log\tfrac{1}{\varepsilon}\big).
\]
(ii) If $u=\psi(c,h_1,\dots,h_P)$ reads the states $h_i$ of $P$ controller tokens (the listener) and a mask $M$ routes rule information across image tokens through those tokens, then
\[
\mathcal{R}_{\mathrm{rose}}=\{\mathcal{F}_{\mathrm{site}},\mathcal{F}_{\mathrm{ctrl}}\}, \qquad H_\varepsilon(\mathcal{R}_{\mathrm{rose}})=\Theta\big(d\log\tfrac{1}{\varepsilon}\big).
\]
(iii) If $\psi$ returns $u=c$ at a zero gate and the structural pass reuses the dense weights, then $\mathcal{F}_{\mathrm{dense}}\subseteq\mathcal{F}_{\mathrm{rose}}$ at the same level of parameter count, where $\mathcal{F}_{\mathrm{dense}}$ and $\mathcal{F}_{\mathrm{rose}}$ are the full function classes of the dense backbone and \rose{}.

\textit{Proof sketch.} Clause (ii) is RoSeViT's theorem with $u$ as the rule variable, since every image token's modulation is one function of the $d$-dimensional state $u$ and the mask is the enforcement.
Clause (i) is its collapse condition because the modulation hands every token an independent site-wise rule state, and (iii) holds by construction.

The theorem points out that the adaLN of DiT and SiT computes a fixed $u=c$, which collapses the rule conditioning.
Thus the controller route itself cannot directly affect the fixed adaLN computation, and this causes the bottleneck.
The only way to address this issue is to introduce the listening mechanism, which makes adaLN compute under the controller's command.

\rose{} is the realization of clauses (ii) and (iii).
It builds $P=12$ controller tokens from the condition, reuses the dense weights in the structural pass, and lets every block's modulation read the controller state through $\psi$ (\cref{sec:method}).
By the theorem it removes the bottleneck at the same level of parameter count.
The entropy theorem predicts our \rose{} would outperform SiT under the same level of parameter count, and the gain would partially collapse if the second-pass enforcement were removed, as witnessed in \cref{sec:ablation}.

\section{Role-Separated Diffusion Transformer}
\label{sec:method}

\begin{figure*}[t]
\centering
\includegraphics[width=0.95\textwidth]{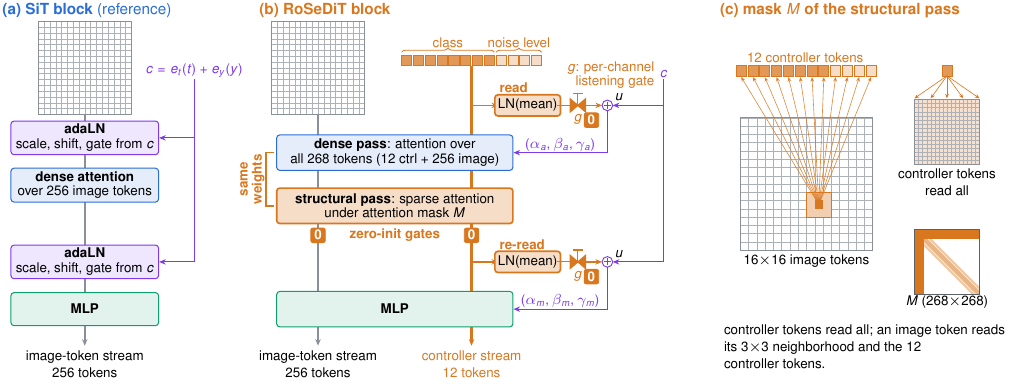}
\caption{\textbf{The \rose{} block.} Twelve controller tokens are prepended to the image tokens. Each block combines dense attention over all tokens with a structural pass that restricts each image token to its $3{\times}3$ spatial neighborhood and the controller tokens, while allowing controller tokens to attend globally. The structural pass reuses the dense attention weights. With the listening mechanism, adaLN computes scale, shift, and gating parameters from the fixed condition $c$ and evolving controller states, before dense attention and again before the MLP.}
\label{fig:block}
\end{figure*}

Building on SiT and RoSeViT, \rose{} implements global rule isolation through three complementary mechanisms: (i) dedicated controller tokens that maintain evolving global conditioning representations, (ii) a listening mechanism that incorporates controller states into adaLN modulation, and (iii) a parameter-shared structural pass that promotes controller-mediated information exchange. \Cref{fig:block} illustrates the resulting architecture. Let $x\in\mathbb{R}^{L\times d}$ denote the image tokens and $c=e_t(t)+e_y(y)$ the global conditioning vector, where $e_t$ and $e_y$ are the timestep and class embeddings, respectively.

\subsection{Condition-Initialized Controller Tokens}

\rose{} introduces $P=12$ dedicated controller tokens, comprising eight class-conditioned and four timestep-conditioned tokens. Their initial representations are
\begin{equation}
\begin{aligned}
h^{(0)}_k &= b^{(y)}_k+e_y(y), & k&=1,\dots,8,\\
h^{(0)}_{8+k} &= b^{(t)}_k+e_t(t), & k&=1,\dots,4,
\end{aligned}
\label{eq:ctrl}
\end{equation}
where each $b^{(y)}_k,b^{(t)}_k\in\mathbb{R}^d$ is a learned token embedding. The controller tokens reuse SiT's class and timestep embeddings, introducing no additional conditioning inputs. The fixed conditioning vector $c$ is retained for adaLN, while the controller tokens provide evolving representations of the same global condition. The token allocation follows iMF~\cite{geng2025imf}.

The controller tokens are prepended to the image tokens without additional positional embeddings:
\[
z^{(0)}=[h^{(0)}_1,\dots,h^{(0)}_P,\,x_1,\dots,x_L]
\in\mathbb{R}^{(P+L)\times d}.
\]
Unlike the fixed conditioning vector $c$, the controller tokens maintain their own residual streams and are updated across transformer blocks through interactions with image tokens. Thus, their states $h_i^{(l)}$ progressively incorporate image-dependent information while retaining the initial conditioning context. At the first block, the controller tokens contain only the initialized condition; subsequent blocks can use information accumulated through attention.

Using multiple controller tokens provides a compact set of evolving representations for class and timestep conditioning rather than a single fixed vector. Their states influence image-token processing through the adaLN listening mechanism and the structural pass, as detailed in the following subsections. Only image-token outputs are unpatchified to predict the diffusion velocity.

\subsection{The adaLN Listening Mechanism}
\label{sec:listening}

The listening mechanism enables adaLN modulation to incorporate evolving controller states rather than relying solely on fixed condition embeddings. At each block, the conditioning representation is computed as
\begin{equation}
u = c + g\odot \mathrm{LN}\!\left(\frac{1}{P}\sum_{i=1}^{P}h_i\right),
\label{eq:listen}
\end{equation}
where $c=e_t(t)+e_y(y)$ is the original conditioning vector, $h_i$ are the current controller states, and $g\in\mathbb{R}^d$ is a learned per-channel listening gate. The adaLN scale, shift, and residual-gating parameters for attention and the MLP are then obtained as $(\alpha_a,\beta_a,\gamma_a,\alpha_m,\beta_m,\gamma_m)=Wu$. These modulations are applied to image tokens, allowing controller information to influence their processing.

Each block reads the controller states twice: before dense attention, using the states available at the block entrance, and after the structural pass, incorporating the updated controller representations. The first read determines attention modulation, while the second determines MLP modulation. The final output layer uses a separate listening gate. Unlike SiT, whose adaLN depends only on the fixed condition $c$, RoSeDiT dynamically couples conditioning to the evolving image representations through the controller tokens. This connection is essential because controller tokens alone do not ensure that their information influences image-token modulation.

The listening gates and additional residual gates are zero-initialized, initially suppressing controller-dependent modulation and the structural pass contribution. The original conditioning pathway is retained, allowing the controller pathway to be learned progressively. Controller embeddings, listening gates, and residual gates together increase the parameter count by less than $0.1\%$ relative to SiT.

\subsection{Parameter-Shared Structural Pass}
\label{sec:structural}

Following RoSeViT, RoSeDiT introduces a structural pass after dense attention to reinforce controller-mediated information exchange. Its attention mask is defined as
\[
M_{ij}=1 \iff i\le P \;\text{or}\; j\le P
\;\text{or}\; j\in\mathcal{N}_{3\times3}(i),
\]
where $P=12$ and $\mathcal{N}_{3\times3}(i)$ denotes the spatial neighborhood of image token $i$. Controller tokens attend globally, while each image token attends only to its local $3\times3$ neighborhood and the controller tokens within this pass. Thus, the structural pass provides a constrained pathway in which nonlocal image information is communicated through controller tokens rather than unrestricted image-to-image attention.

The structural pass reuses the query, key, value, and output projections of dense attention, introducing no additional attention parameters. Its contribution enters the residual stream through a zero-initialized gate. Importantly, the original dense attention pass is retained, preserving direct global image-token interactions alongside the controller-mediated pathway. Together with the adaLN listening mechanism, the structural pass encourages the model to aggregate global conditioning information in the controller tokens and use it to modulate spatial representations.

\section{Deterministic Sampling Manifold Compression}
\label{sec:dsmc}

\begin{figure}[t]
\centering
\includegraphics[width=0.95\linewidth]{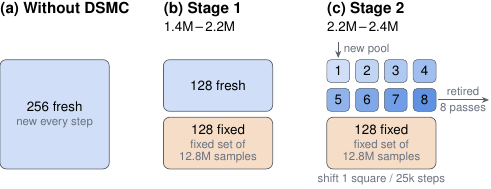}
\caption{\textbf{Deterministic Sampling Manifold Compression (DSMC).} (a) Standard training samples fresh (image, noise level, noise) triples. (b) Stage 1 combines 128 fresh triples with 128 triples from a fixed set $F$ in each batch of 256. (c) Stage 2 retains the fixed triples while replacing the fresh half with eight rotating subsets of 16 triples, each drawn from a pool reused for eight passes. A new pool enters every 25K steps.}
\label{fig:dsmc}
\end{figure}

We introduce Deterministic Sampling Manifold Compression (DSMC), a two-stage training schedule for the continued training of diffusion models. DSMC combines deterministic reuse of (image, noise level, noise) triples with stochastic sampling. It changes which triples a batch contains and nothing else.

\paragraph{Observed phenomenon.}
Let $\mathcal{L}(\theta)$ denote the denoising objective of a model $\theta$ on fresh triples. Since $\mathcal{L}$ is the training objective, it is generally understood to measure how well the image manifold is learned, so that a better manifold implies a lower loss:
\begin{equation}
\mathrm{FID}(\theta_B)<\mathrm{FID}(\theta_A)\;\Rightarrow\;\mathcal{L}(\theta_B)<\mathcal{L}(\theta_A).
\label{eq:dsmc-assumption}
\end{equation}
We observed that this does not hold for two arms A and B continued from the same checkpoint. Arm A receives only fresh triples, and arm B receives half fresh triples and half triples from a fixed set. After 800K steps, the losses on fresh triples are the same, $\mathcal{L}(\theta_A)=0.736$ and $\mathcal{L}(\theta_B)=0.738$. But B reaches an FID of 6.75, while A needs 200K more steps to reach 7.61. This FID gap at the same loss indicates a misalignment between the denoising objective and the image manifold. DSMC formalizes this phenomenon into a two-stage schedule that reuses identical triples, deterministically generated from fixed seeds and sample indices, alongside fresh triples.

\paragraph{Stage 1: Fixed-sample reuse.}
From the 1.4M-step checkpoint to 2.2M steps, each batch contains 128 freshly sampled triples and 128 triples drawn from a fixed set $F$ of 12.8M triples (\cref{fig:dsmc}b). The fixed triples are traversed in shuffled order, completing one pass every 100K steps and eight passes over this stage. This combines repeated optimization on fixed targets with continued exposure to fresh triples.

\paragraph{Stage 2: Rotating sample pools.}
From 2.2M to 2.4M steps, the fixed half continues traversing $F$ for two additional passes. The remaining 128 triples are divided into eight groups of 16 (\cref{fig:dsmc}c). Each group draws from a pool of 400K fixed triples, completing one pass every 25K steps. A new pool enters the first group every 25K steps, while existing pools advance through successive groups and are retired after eight passes, so the first group always holds a pool on its first pass. This staggered schedule repeatedly optimizes individual pools while progressively introducing new triples. All other training settings, including batch size and learning rate, remain unchanged, and the schedule ends at 2.4M steps.

\section{Experiments}
\label{sec:experiments}

We evaluate \rose{} against DiT and SiT across four model scales (\cref{sec:ladder}), examine continued training with DSMC (\cref{sec:longer}), assess classifier-free guidance (\cref{sec:cfg}) and compatibility with REPA (\cref{sec:repa}), and conduct architectural ablations (\cref{sec:ablation}) to investigate the contributions of listening, the structural pass, and the role-separation principle.

\subsection{Experimental Setup}
\label{sec:setup}

We train class-conditional diffusion transformers on ImageNet~\cite{deng2009imagenet} at $256{\times}256$ resolution using the latent space of the Stable Diffusion autoencoder~\cite{rombach2022ldm} and the official REPA implementation~\cite{yu2025repa}. Following the SiT training protocol, we use linear interpolation with velocity prediction, AdamW~\cite{loshchilov2019adamw} with a constant learning rate of $10^{-4}$ and zero weight decay, a batch size of 256, and an exponential moving average (EMA) of model parameters with decay 0.9999. Unless otherwise specified, models are trained for 400K steps.

We evaluate 50,000 generated samples from EMA models using the ADM evaluation suite~\cite{dhariwal2021adm}, reporting FID~\cite{heusel2017fid}, spatial FID (sFID)~\cite{nash2021sfid}, Inception Score (IS)~\cite{salimans2016is}, and precision and recall~\cite{kynkaanniemi2019pr}. Samples are generated using a 250-step Euler--Maruyama SDE sampler. All primary results are reported without classifier-free guidance (CFG); guided results and their corresponding guidance scales are presented separately in \cref{sec:cfg}.

We compare against published DiT, SiT, and SiT + REPA results under their reported evaluation protocols. For controlled ablations, we additionally reproduce SiT-B/2 using our training pipeline, obtaining an FID of 35.54 at 400K steps. We distinguish this reproduced baseline from the published SiT results.

\subsection{Main Results}
\label{sec:ladder}

\begin{table}[t]
\centering\footnotesize
\setlength{\tabcolsep}{3pt}
\caption{
\textbf{Main results on ImageNet $256{\times}256$ generation} without classifier-free guidance, evaluated on 50K samples. (a) At 400K steps, \rose{} outperforms DiT and SiT across four model scales, matching SiT in depth $N$, width $d$, and training protocol at the same level of parameter count. The B/2 rows below the separator compare SiT and \rose{} with REPA using DINOv2-B alignment. (b) Longer-training comparisons: \rose{}-XL/2 achieves an FID of 7.98 at 1.4M steps, surpassing SiT-XL/2's 8.3 at 7M steps. With DSMC, \rose{}-XL/2 reaches 5.84 at 2.4M steps, outperforming SiT-XL/2 + REPA's 5.9 at 4M steps. $^\dagger$Our reproduction.
}
\label{tab:ladder}
\textbf{(a) 400K steps, four model scales}\par\smallskip
\resizebox{\linewidth}{!}{%
\begin{tabular}{lrrr|ccccc}
\toprule
Model & $N$ & $d$ & params & FID$\downarrow$ & sFID$\downarrow$ & IS$\uparrow$ & Prec.$\uparrow$ & Rec.$\uparrow$ \\
\midrule
DiT-S/2      & 12 & 384  & 33M  & 68.4 & -- & -- & -- & -- \\
SiT-S/2      & 12 & 384  & 33M  & 57.6 & -- & -- & -- & -- \\
\rose{}-S/2  & 12 & 384  & 33M  & \textbf{51.42} & 8.69 & 28.7 & .433 & .618 \\
\midrule
DiT-B/2      & 12 & 768  & 130M & 43.5 & -- & -- & -- & -- \\
SiT-B/2      & 12 & 768  & 130M & 33.0 & -- & -- & -- & -- \\
SiT-B/2$^\dagger$ & 12 & 768 & 130M & 35.54 & 6.52 & 42.0 & .531 & .636 \\
\rose{}-B/2  & 12 & 768  & 130M & \textbf{27.18} & \textbf{6.26} & \textbf{56.5} & \textbf{.568} & \textbf{.638} \\
\cmidrule(lr){1-9}
SiT-B/2 + REPA & 12 & 768 & 130M & 24.4 & -- & -- & -- & -- \\
\rose{}-B/2 + REPA & 12 & 768 & 130M & \textbf{21.01} & 6.23 & 68.3 & .606 & .649 \\
\midrule
DiT-L/2      & 24 & 1024 & 458M & 23.3 & -- & -- & -- & -- \\
SiT-L/2      & 24 & 1024 & 458M & 18.8 & -- & -- & -- & -- \\
\rose{}-L/2  & 24 & 1024 & 458M & \textbf{15.34} & 5.16 & 84.6 & .654 & .636 \\
\midrule
DiT-XL/2     & 28 & 1152 & 675M & 19.5 & -- & -- & -- & -- \\
SiT-XL/2     & 28 & 1152 & 675M & 17.2 & -- & -- & -- & -- \\
\rose{}-XL/2 & 28 & 1152 & 675M & \textbf{12.63} & 5.06 & 97.0 & .678 & .634 \\
\bottomrule
\end{tabular}}
\par\medskip
\textbf{(b) Longer-training comparison}\par\smallskip
\resizebox{\linewidth}{!}{%
\begin{tabular}{lrr|ccccc}
\toprule
Model & steps & epochs & FID$\downarrow$ & sFID$\downarrow$ & IS$\uparrow$ & Prec.$\uparrow$ & Rec.$\uparrow$ \\
\midrule
DiT-XL/2            & 7M   & 1400 & 9.6  & 6.85 & 121.5 & .67  & .67  \\
SiT-XL/2            & 7M   & 1400 & 8.3  & 6.32 & 131.7 & .68  & .67  \\
\rose{}-XL/2        & 1.4M & 280  & 7.98 & \textbf{5.30} & 129.1 & \textbf{.696} & .660 \\
\rose{}-XL/2        & 2.4M & 480  & \textbf{7.61} & 5.49 & \textbf{137.0} & .694 & \textbf{.670} \\
\midrule
SiT-XL/2 + REPA     & 4M   & 800  & 5.9  & 5.73 & \textbf{157.8} & .70  & \textbf{.69}  \\
\rose{}-XL/2 + DSMC & 2.4M & 480  & \textbf{5.84} & \textbf{4.67} & 150.4 & \textbf{.708} & .666 \\
\bottomrule
\end{tabular}}
\end{table}

\begin{figure}[t]
\centering
\includegraphics[width=0.9\linewidth]{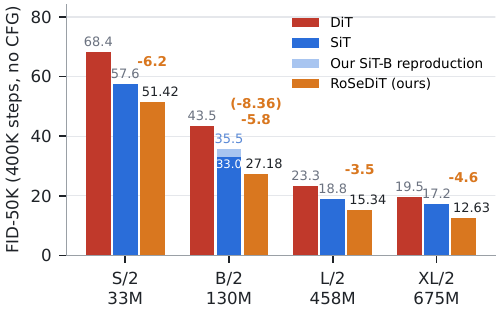}
\caption{
\textbf{FID comparison across model scales at 400K steps.} \rose{} consistently outperforms DiT and SiT at all four model scales under matched training budgets. Orange annotations indicate  FID improvement relative to SiT at each scale.}
\label{fig:ladder}
\end{figure}

\Cref{tab:ladder,fig:ladder} compare \rose{} with DiT and SiT across four model scales at 400K steps. Under matched model configurations and training protocols, \rose{} consistently outperforms SiT: FID improves from 57.6 to 51.42 at S/2, 33.0 to 27.18 at B/2, 18.8 to 15.34 at L/2, and 17.2 to 12.63 at XL/2. The relative FID reduction increases from 11\% at S/2 to 27\% at XL/2, suggesting that the benefits of role-separated conditioning extend across model scales. \rose{} also outperforms DiT at every corresponding scale.

Beyond matched-size comparisons, \rose{} demonstrates favorable performance relative to model capacity. \rose{}-L/2 achieves an FID of 15.34 with 458M parameters, outperforming SiT-XL/2's 17.2 with 675M parameters. This represents a 1.9-point FID improvement with 32\% fewer parameters.

The advantage persists with longer training (\cref{fig:track}). Without changing the training recipe, \rose{}-XL/2 improves from an FID of 12.63 at 400K steps to 7.98 at 1.4M steps, surpassing SiT-XL/2's FID of 8.3 after 7M steps. Thus, \rose{} achieves better FID using one-fifth as many training steps (280 versus 1400 epochs), indicating substantial training-step efficiency. Additional parameter-count controls in \cref{sec:ablation} support that these improvements arise from the architecture rather than increased model capacity.

\subsection{Continued Training with DSMC}
\label{sec:longer}
\begin{figure}[t]
\centering
\includegraphics[width=0.9\linewidth]{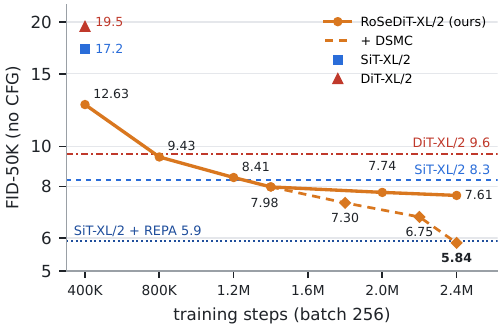}
\caption{
\textbf{Training progression of \rose{}-XL/2 on ImageNet $256{\times}256$ without classifier-free guidance.} Published DiT, SiT, and SiT + REPA results are shown for reference. \rose{}-XL/2 reaches an FID of 7.98 at 1.4M steps, outperforming SiT-XL/2's 8.3 at 7M steps. Starting from the same 1.4M-step checkpoint, DSMC (dashed curve) achieves an FID of 5.84 at 2.4M steps, compared with 7.61 under standard continued training, and surpasses SiT-XL/2 + REPA's 5.9 at 4M steps.}
\label{fig:track}
\end{figure}

To evaluate DSMC, we continue training \rose{}-XL/2 from the same 1.4M-step checkpoint (FID 7.98) for an additional 1M steps, with and without DSMC (\cref{fig:track}). Under standard training, FID improves to 7.74 at 2.0M steps and 7.61 at 2.4M steps. With DSMC (\cref{sec:dsmc}), FID reaches 6.75 at 2.2M steps after Stage 1 and 5.84 at 2.4M steps after Stage 2. At the same training budget of 2.4M steps, DSMC improves FID from 7.61 to 5.84, a reduction of 1.77 points, while keeping the architecture, batch size, learning rate, and other training settings unchanged. The DSMC trajectory continues improving through Stage 1 without an observed plateau, and continuing it may lead to a better result that we didn't test.

With DSMC, \rose{}-XL/2 achieves an FID of 5.84 after 2.4M steps (480 epochs), outperforming SiT-XL/2 + REPA's 5.9 after 4M steps (800 epochs), without representation alignment. Unlike REPA, DSMC requires no pretrained teacher, distillation objective, or architectural modification. These results demonstrate that deterministic sample reuse provides a complementary means of improving diffusion generation quality with fewer training steps.

\subsection{Classifier-Free Guidance}
\label{sec:cfg}
\begin{table}[t]
\centering\scriptsize
\setlength{\tabcolsep}{2pt}
\caption{
\textbf{Classifier-free guidance results} on 50K samples using the SDE sampler. CFG specifies the guidance scale and optional noise-level interval. \textbf{Top:} \rose{} across four model scales at 400K steps. \textbf{Middle:} B/2 architecture and capacity comparisons. \textbf{Bottom:} Longer-trained XL/2 models and ImageNet training/validation self-evaluation. $^\dagger$Our reproduction.
}
\label{tab:cfg}
\resizebox{\linewidth}{!}{%
\begin{tabular}{lrrl|ccccc}
\toprule
Model & params & steps & CFG & FID$\downarrow$ & sFID$\downarrow$ & IS$\uparrow$ & Prec.$\uparrow$ & Rec.$\uparrow$ \\
\midrule
\rose{}-S/2  & 33M  & 400K & 1.5 & \textbf{23.31} & 6.42 & 68.6  & .610 & .542 \\
\rose{}-B/2  & 130M & 400K & 1.5 & \textbf{7.15} & 5.05 & 153.2 & .775 & .500 \\
\rose{}-L/2  & 458M & 400K & 1.5 & \textbf{3.88} & 4.85 & 213.4 & .840 & .497 \\
\rose{}-XL/2 & 675M & 400K & 1.5 & \textbf{3.66} & 4.92 & 230.6 & .852 & .496 \\
\midrule
SiT-B/2$^\dagger$ & 130M & 400K & 1.5 & 12.23 & 5.23 & 108.8 & .729 & \textbf{.516} \\
\quad depth 15$^\dagger$ & 162M & 400K & 1.5 & 10.10 & 5.09 & 121.2 & .753 & .515 \\
\quad width 864$^\dagger$ & 165M & 400K & 1.5 & 11.18 & 5.13 & 115.8 & .737 & .512 \\
\rose{}-B/2 & 130M & 400K & 1.5 & \textbf{7.15} & 5.05 & \textbf{153.2} & .775 & .500 \\
\rose{}-Dualweight-B/2 & 160M & 400K & 1.5 & 8.09 & \textbf{4.88} & 136.8 & \textbf{.783} & .500 \\
\midrule
DiT-XL/2 & 675M & 7M & 1.5 & 2.27 & 4.60 & 278.2 & .83 & .57 \\
SiT-XL/2 & 675M & 7M & 1.5 & 2.06 & 4.49 & 277.5 & \textbf{.83} & .59 \\
\rose{}-XL/2 & 675M & 1.4M & 1.8 [0,0.7] & 1.95 & 4.53 & 290.9 & \textbf{.83} & .586 \\
\rose{}-XL/2 & 675M & 2.4M & 1.8 [0,0.7] & 1.76 & 4.57 & \textbf{301.5} & .820 & .605 \\
\rose{}-XL/2 + DSMC & 675M & 2.4M & 1.5 [0,0.7] & \textbf{1.70} & \textbf{4.38} & 254.2 & .791 & \textbf{.636} \\
\midrule
ImageNet train on train & \multicolumn{1}{c}{--} & \multicolumn{1}{c}{--} & \multicolumn{1}{c|}{--} & 0.85 & 3.42 & 336.1 & .756 & .719 \\
ImageNet train on val   & \multicolumn{1}{c}{--} & \multicolumn{1}{c}{--} & \multicolumn{1}{c|}{--} & 2.23 & 6.79 & 336.1 & .670 & .706 \\
ImageNet val on train   & \multicolumn{1}{c}{--} & \multicolumn{1}{c}{--} & \multicolumn{1}{c|}{--} & 1.68 & 3.67 & 232.4 & .752 & .667 \\
\bottomrule
\end{tabular}}
\end{table}

We observe that ImageNet self-evaluation is inconsistent across its training and validation sets (\cref{tab:cfg}, bottom). A major performance shift happens across all five metrics. On the FID side, the score shifts from 0.85 to 2.23 to 1.68 by switching the sides of the training and validation sets. Therefore, any FID score below 2.23 could be misleading and questionable. We make no claim on the flagship CFG scores and list them as a secondary result.

At CFG scale 1.5 and 400K steps, \rose{}-L/2 and \rose{}-XL/2 achieve FIDs of 3.88 and 3.66. At B/2, \rose{} improves the FID from our reproduced SiT-B/2's 12.23 to 7.15. For \rose{}-Dualweight, we create two additional parameter-matched SiT baselines, SiT Depth 15 and SiT Width 864. \rose{}-Dualweight reaches FID 8.09 and surpasses the two baselines' 10.10 and 11.18.

Under the same level of parameter count, \rose{}-B/2 drastically surpasses SiT-B/2 by 5.08 FID. Compared with the two additional depth and width variants, \rose{}-B/2 holds its advantage by 2.95 and 4.03 FID with 32M and 34M fewer parameters. \rose{}-Dualweight also surpasses the two baselines by 2.01 and 3.09 FID. Interestingly, this dual-weight version didn't surpass \rose{} with 30M more parameters, which shows the structural benefit is not from the parameter count.

\begin{table}[t]
\centering\footnotesize
\setlength{\tabcolsep}{4pt}
\caption{
\textbf{Ablations at B/2, 400K steps.} \textbf{Top:} RoSeViT-style structural pass with fixed modulation, rank-64 listening, and the final \rose{} architecture. \textbf{Middle:} Single-pass listening with in-context controller tokens and independent SiT depth/width controls. \textbf{Bottom:} \rose{}-Dualweight with pooled controller tokens and fixed modulation, ablating the attention mask and controller tokens. CFG uses scale 1.5. Bold indicates the best result per group. $^\dagger$Our reproduction.
}
\label{tab:ablation}
\begin{tabular}{lrccc}
\toprule
Model (B/2) & Params & FID$\downarrow$ & sFID$\downarrow$ & FID (cfg 1.5) $\downarrow$\\
\midrule
SiT-B/2$^\dagger$ & 130M & 35.54 & 6.52 & 12.23 \\
RoSeViT  (fixed adaLN) & 130M & 34.03 & -- & 11.75 \\
\quad + Listening (rank-64) & 132M & 27.92 & 6.28 & 7.39 \\
\textbf{\rose{}-B/2} & 130M & \textbf{27.18} & \textbf{6.26} & \textbf{7.15} \\
\midrule
SiT-B/2$^\dagger$ & 130M & 35.54 & 6.52 & 12.23 \\
\quad + In-context controller & 130M & 34.17 & 6.38 & 11.48 \\
\quad \quad + Listening (rank-64) & 132M & \textbf{30.26} & 6.52 & \textbf{8.19} \\
SiT-B/2 Depth (15)$^\dagger$ & 162M & 31.65 & \textbf{6.20} & 10.10 \\
SiT-B/2 Width (864)$^\dagger$ & 165M & 33.83 & 6.32 & 11.18 \\
\midrule
\rose{}-Dualweight-B/2 & 160M & \textbf{26.44} & \textbf{5.75} & \textbf{8.09} \\
\quad - Mask & 160M & 27.98 & 5.85 & 8.89 \\
\quad - Controller & 159M & 27.44 & 5.90 & 8.33 \\
\quad - Both & 159M & 29.07 & 6.07 & 9.33 \\
\bottomrule
\end{tabular}
\end{table}

\subsection{Compatibility with REPA}
\label{sec:repa}

REPA~\cite{yu2025repa} improves diffusion training through representation alignment without requiring changes to the transformer block, making it complementary to \rose{}. We combine \rose{}-B/2 with REPA using its default configuration and train for 400K steps (\cref{tab:ladder}a). REPA improves \rose{}-B/2's FID from 27.18 to 21.0. Under the same REPA setting, \rose{} outperforms SiT-B/2 (24.4) by 3.4 FID and achieves a guided FID of 5.29. These results demonstrate that the benefits of role-separated conditioning persist when combined with representation alignment.

\subsection{Ablations}
\label{sec:ablation}

\Cref{tab:ablation} reports three groups of ablations at B/2 against our reproduced SiT-B/2 (FID 35.54). The first group builds \rose{} from SiT step by step, and shows which component brings the gain. The second group adds the controller and the listening mechanism to SiT without the structural pass, and shows that SiT benefits from listening as well. The third group ablates \rose{}-Dualweight, which gives the structural pass its own weights, and shows the role of the second-pass enforcement and of the parameter count.

\paragraph{From SiT to \rose{}.}
Directly transferring RoSeViT with image-pooled controller tokens, a shared structural pass, and fixed adaLN yields only 34.03, a minor improvement of 1.51 FID at the same level of parameter count. This is the case predicted in \cref{sec:background}, where the controller route cannot affect a fixed adaLN. Enabling listening with a rank-64 linear readout (+1.3M params) makes the FID jump to 27.92, another 6.11. The final \rose{} architecture replaces this readout with the per-channel diagonal listening gate and reaches 27.18. With fewer parameters, the diagonal gate surpasses the rank-64 linear readout by 0.74 FID. At the same level of parameter count, \rose{}-B/2 surpasses SiT-B/2 by 8.36 FID, and the listening mechanism brings most of this gain.

\paragraph{SiT with listening.}
SiT shows a similar benefit from the listening mechanism. By adding the in-context controller alone, SiT shows the same trend as \rose{}, with only a minor improvement from 35.54 to 34.17. On top of that, adding the listening mechanism with the same rank-64 linear readout makes SiT's performance jump to an FID of 30.26. It surpasses the two larger variants, SiT Depth 15 at 31.65 and SiT Width 864 at 33.83, with 31M and 33M fewer parameters. It still doesn't match \rose{}-B/2 at 27.18, due to the lack of the second-pass enforcement. With the same readout, adding the structural pass improves the FID from 30.26 to 27.92.

\paragraph{\rose{}-Dualweight.}
\rose{}-Dualweight gives the structural pass its own weights with 30M more parameters, and reaches 26.44. At the same level of parameter count, it surpasses SiT Depth 15 at 31.65 and SiT Width 864 at 33.83 by 5.21 and 7.39 FID. Removing the attention mask increases its FID to 27.98, removing the controller tokens yields 27.44, and removing both gives 29.07. The gain partially collapses once the second-pass enforcement is removed, as predicted in \cref{sec:background}. Compared with \rose{}-B/2, \rose{}-Dualweight is 0.74 FID better without guidance and 0.94 worse with guidance. Its 30M more parameters show no significant advantage.

\section{Conclusion}
\label{sec:conclusion}

We presented \rose{}, the Role-Separated Diffusion Transformer. By introducing the adaLN listening mechanism, \rose{} brings the entropy theorem and role separation into the latent diffusion domain. At the same level of parameter count, \rose{} surpasses SiT at all four model scales, and \rose{}-XL/2 surpasses SiT-XL/2's 7M-step FID of 8.3 at 1.4M steps with 7.98. With DSMC, a training schedule that leaves the model unchanged, \rose{}-XL/2 reaches 5.84 at 2.4M steps and surpasses SiT + REPA's 5.9 at 4M steps. These results demonstrate the potential of role-separated conditioning and deterministic sample reuse for diffusion transformer training.

\section*{LLM Usage Statement}
Claude Code (Anthropic) was used in the experiments and in the writing of this paper. In the experiments, it was used to write and edit code, to run and monitor training and evaluation, and to tabulate results. In the writing, it was used to draft and edit text under the authors' direction, to draw the schematic figures from the authors' specifications, and to check bibliographic entries. ChatGPT (OpenAI) and Claude were also used to obtain feedback on drafts of the manuscript. The method of this work was designed by the authors. Claude Code did not contribute to the research idea, the method design, or the reading of the data. The authors reviewed all content and take full responsibility for it.
\end{document}